\documentclass[preprint,twocolumn,3p,authoryear]{elsarticle}
\usepackage{amssymb}  
\usepackage{ucs}
\usepackage{comment}
\usepackage{amsmath}
\usepackage{pbox}
\usepackage{tcolorbox}
\usepackage{caption}
\usepackage{capt-of}
\usepackage{bm}
\usepackage{graphicx}

\usepackage{array}
\usepackage{multirow}
\usepackage{colortbl}
\usepackage{tikz}
\usepackage{xspace}
\usepackage{hyperref}

\renewcommand{\cite}{\citep}

\newcommand{\vips}{{\sc vips}\xspace}
\newcommand{\reps}{{\sc reps}\xspace}

\newcommand{\aaac}{{\sc a3c}\xspace}

\newcommand{\reinforce}{{\sc reinforce}\xspace}

\newcommand{\rock}{{\sc rock}$^*$\xspace}
\newcommand{\ddpg}{{\sc ddpg}\xspace}
\newcommand{\dfpg}{{\sc d4pg}\xspace}
\newcommand{\sac}{{\sc sac}\xspace}
\newcommand{\tddd}{{\sc td3}\xspace}
\newcommand{\acktr}{{\sc acktr}\xspace}
\newcommand{\naf}{{\sc naf}\xspace}
\newcommand{\qprop}{{\sc Q-prop}\xspace}
\newcommand{\acer}{{\sc acer}\xspace}
\newcommand{\pgql}{{\sc pgql}\xspace}
\newcommand{\ipg}{{\sc ipg}\xspace}
\newcommand{\cmaes}{{\sc cma-es}\xspace}

\newcommand{\nes}{{\sc nes}\xspace}
\newcommand{\xnes}{x{\sc nes}\xspace}
\newcommand{\pepg}{{\sc pepg}\xspace}

\newcommand{\gpomdp}{{\sc g(po)mdp}\xspace}

\newcommand\cem{\textsc{cem}\xspace}
\newcommand{\trpo}{{\sc trpo}\xspace}
\newcommand{\ppo}{{\sc ppo}\xspace}
\newcommand{\gps}{{\sc gps}\xspace}
\newcommand{\nac}{{\sc nac}\xspace}
\newcommand{\enac}{e{\sc nac}\xspace}
\newcommand{\inac}{i{\sc nac}\xspace}

\newcommand{\power}{{\sc p}o{\sc wer}\xspace}

\newcommand{\PIBB}{{\sc pi$^{\mbox{\tiny BB}}$}\xspace}
\newcommand{\picma}{{\sc pi$^2$-cma}\xspace}
\newcommand{\pis}{{\sc pi$^{2}$}\xspace}

\newcommand{\dmps}{{\sc dmp}s\xspace}

\newcommand{\rl}{RL\xspace}
\newcommand{\drl}{deep RL\xspace}
\newcommand{\Drl}{Deep RL\xspace}
\newcommand{\dqn}{{\sc dqn}\xspace}

\newcommand{\ofl}{off-line\xspace}

\newcommand{\ofp}{off-policy\xspace}
\newcommand{\onp}{on-policy\xspace}
\newcommand{\td}{temporal difference\xspace}

\newcommand{\edas}{{\sc EDA}s\xspace}
\newcommand{\pse}{policy search\xspace}

\newcommand{\vgo}{vanilla gradient optimization\xspace}

\newcommand{\mfo}{optimization without a utility model\xspace}

\newcommand{\roll}{rollout\xspace} 
\newcommand{\rolls}{rollouts\xspace} 
\newcommand{\gep}{\textsc{gep}\xspace}
\newcommand{\geps}{\textsc{gep}s\xspace}

\newcommand{\sarsa}{{\sc Sarsa}\xspace}
\newcommand{\ql}{{\sc q-learning}\xspace}

\newcommand{\mymath}[1]{\ensuremath{#1}\xspace}
\newcommand{\mymathbf}[1]{\mymath{\mathbf{\boldsymbol{#1}}}}
\newcommand{\covar}{\mymathbf{\Sigma}}

\renewcommand{\vec}[1]{\mymathbf{#1}}

\newcommand{\vu}{\vec{u}}
\newcommand{\vx}{\vec{x}}

\newcommand{\U}{\mathcal{U}}
\newcommand{\X}{\mathcal{X}}

\newcommand{\hatu}{\mymath{\hat{U}}}

\newcommand\wi[1]{$\circ$}
\newcommand\bu[1]{$\bullet$}
\newcommand\ot[1]{$\star$}
\newcommand\spa[1]{$\spadesuit$}
\newcommand\dn[1]{.}

\newcommand\pol{\pi} 
\newcommand{\params}{{\mymathbf{\theta}}\xspace} 
\newcommand{\Params}{{\mymathbf{\Theta}}\xspace}

\newcounter{cbox} 
\newcommand{\cbox}{\arabic{cbox}}

\newcounter{cmes} 
\newcommand{\cmes}{\arabic{cmes}}
\newenvironment{mymes}[1]{\begin{tcolorbox}[colback=red!10!white]\refstepcounter{cmes}{\bf Message \cmes:}\label{#1}}{\end{tcolorbox}}

\newcounter{csum} 
\newcommand{\csum}{\arabic{csum}}

\definecolor{myred}{rgb}{0.8,0,0}
\definecolor{mygreen}{rgb}{0,0.6,0}
\definecolor{myblue}{rgb}{0,0,0.7}
\definecolor{mygray}{rgb}{0.4,0.4,0.4}
\definecolor{mygreen2}{rgb}{0.0,0.3,0.0}

\usepackage[textsize=tiny]{todonotes}

\newcommand{\new}[1]{#1}
\newcommand{\removed}[1]{}
\newcommand{\rephrased}[1]{#1}

\title{Policy Search in Continuous Action Domains: an Overview}

\graphicspath{{./}}

\begin{document}
\begin{frontmatter}

  \author{Olivier Sigaud}
  \address{INRIA Bordeaux Sud-Ouest, \'equipe FLOWERS\\
    Sorbonne Universit\'e, CNRS UMR 7222,\\
    Institut des Syst\`emes Intelligents et de Robotique, F-75005 Paris, France\\
    {\tt olivier.sigaud@isir.upmc.fr}~~~~+33 (0) 1 44 27 88 53
  }

  \author{Freek Stulp}
  \address{German Aerospace Center (DLR), Institute of Robotics and Mechatronics, Wessling, Germany\\
    {\tt freek.stulp@dlr.de}
  }

  \begin{keyword}
    policy search, sample efficiency, deep reinforcement learning, deep neuroevolution
  \end{keyword}

  \begin{abstract}
    Continuous action policy search is currently the focus of intensive research, driven both by the recent success of deep reinforcement learning algorithms and the emergence of competitors based on evolutionary algorithms.
    In this paper, we present a broad survey of \pse methods, providing a unified perspective on very different approaches, including also Bayesian Optimization and directed exploration methods. The main message of this overview is in the relationship between the families of methods, but we also outline some factors underlying sample efficiency properties of the various approaches.
  \end{abstract}
\end{frontmatter}



\section{Introduction}
\label{sec:intro}

Autonomous systems are systems which know what to do in their domain without external intervention.
Generally, their behavior is specified through a {\em policy}.
The policy of a robot, for instance, is defined through a controller which determines actions to take or signals to send to the actuators for any state of the robot in its environment.

\begin{figure}[hbt!]
  \centering
  \includegraphics[width=1.0\columnwidth]{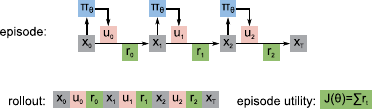}
  \caption{\label{fig:rollout} Visualization of one episode, the information contained in a rollout, and the definition of the episode utility (which is also known as the {\it episode return} when the utility is a reward).}
\end{figure}

Robot policies are often designed by hand, but this manual design is only viable for systems acting in well-structured environments and for well-specified tasks. When those conditions are not met, a more appealing alternative is to let the system find its own policy by exploring various behaviors and exploiting those that perform well with respect to some predefined {\em utility} function. This approach is called {\em \pse}, a particular case of {\em reinforcement learning} (\rl) \cite{sutton98} where
\new{actions are vectors from a continuous space.
More precisely, the goal of \pse is to optimize a policy where the function relating behaviors to their utility is {\it black-box}, i.e. no analytical model or gradient of the utility function is available.}
In practice, a \pse algorithm runs the system with some policies to generate {\em \rolls} made of several state and action {\em steps} and gets the utility as a return (see \figurename~\ref{fig:rollout}).
These utilities are then used to improve the policy, and the process is repeated until some satisfactory set of behaviors is found.
In general, policies are represented with a parametric function, and \pse explores the space of {\em policy parameters}.
For doing so, \roll and utility data are processed by policy improvement algorithms as a set of {\em samples}.

In the context of robotics, {\em sample efficiency} is a key concern.
\new{There are three aspects to sample efficiency}: (1) data efficiency, i.e. extracting more information from available data (definition taken from \cite{deisenroth2011pilco}), (2) sample choice, i.e. obtaining data in which more information is available and (3) sample reuse, i.e. improving a policy several times by using the same samples more than once through {\em experience replay}.

In this paper, we provide a broad overview of \pse algorithms under the perspective of these three aspects.

\subsection{Scope and Contributions}

Three surveys about \pse for robotics have been published in recent years \citep{deisenroth2013survey,stulp13paladyn,kober2013reinforcement}.
With respect to these previous surveys, we cover a broader range of \pse algorithms, including \mfo, Bayesian optimization (BO), directed exploration methods, and \drl.
The counterpart of this breadth is that we do not give a detailed account of the corresponding algorithms nor their mathematical derivation. To compensate for this lack of details, we refer the reader to \cite{deisenroth2013survey} for the mathematical derivation and description of most algorithms before 2013, and we provide carefully chosen references as needed when describing more recent algorithms.

Furthermore, we focus on the case where the system is learning to solve a {\em single task}. That is, we do not cover the broader domain of {\em lifelong}, {\em continual} or {\em open-ended} learning, where a robot must learn how to perform various tasks over a potentially infinite horizon \citep{thrun1995lifelong}.
Additionally, though a subset of \pse methods are based on \rl, we do not cover recent work on \rl with discrete actions such as \dqn and its successors \cite{mnih2015human,hessel2017rainbow}.
Finally, we restrict ourselves to the case where samples are the unique source of information for improving the policy.
That is, we do not consider the interactive context where a human user can provide external guidance \citep{najar2016training},
either through feedback, shaping or demonstration \citep{argall09}.

\subsection{Perspective and structure of the survey}

The main message of this paper is as follows.
In optimization, when the utility function to be optimized is known and convex, efficient convex methods can be applied \citep{gill1981practical}.
If the function is known but not convex, a local optimum can be found using gradient descent, iteratively moving from the current point towards a local optimum by following the direction provided by the derivative of the function at this point.
\rephrased{If the function is \emph{black-box}, neither the function nor its analytic gradient are known.
In policy gradient methods, policy parameters are only related to their utility {\em indirectly} through an intermediate set of observed behaviors.}
Given that \pse corresponds to this more difficult context, we consider five solutions:
\begin{enumerate}
\item
  \label{sol1}
  searching for high utility policy parameters without building a utility model (Section~\ref{sec:mfo}),
\item
  \label{sol2}
  learning a model of the utility function in the space of policy parameters and performing stochastic gradient descent (SGD) using this model (Section~\ref{sec:theta}),
\item
  \label{solb}
  defining an arbitrary {\em outcome space} and using directed exploration \new{of this outcome space} for finding high utility policy parameters (Section~\ref{sec:geps}),
\item
  \label{sol3}
  doing the same as in Solution~\ref{sol2} in the state-action space (Section~\ref{sec:psss}),
\item
  \label{sol4}
  learning a model of the transition function of the system-environment interaction that predicts the next state given the current state and action, to generate samples without using the system, and then applying one of the above solutions based on the generated samples. 
\end{enumerate}
\noindent
An important distinction in the \pse domain is whether the optimization method is episode-based or step-based \citep{deisenroth2013survey}\footnote{This distinction exactly matches the phylogenetic \rl versus ontogenetic \rl distinction in \cite{togelius2009ontogenetic}.}.
The first three solutions above are episode-based, the fourth is step-based and the fifth can be applied to all others.

\begin{figure*}[hbt!]
  \centering
  \includegraphics[width=1.0\textwidth]{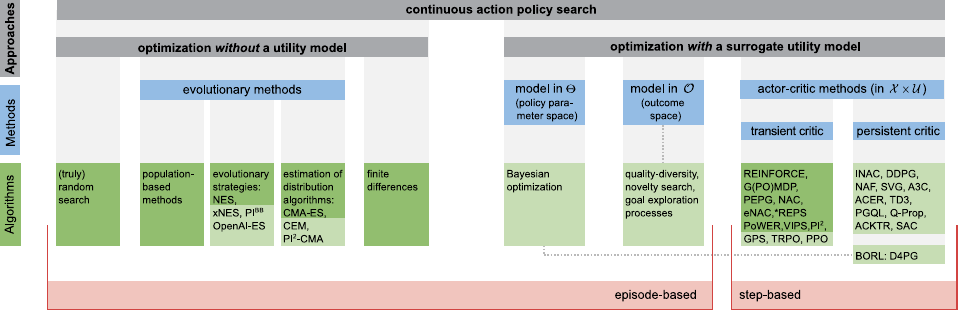}
  \caption{\label{fig:orga_paper} Simplified classification of the algorithms covered in the paper. \Params is the space of policy parameters (see Section~\ref{sec:mfo}), $O$ is an outcome space (see Section~\ref{sec:geps}) and $\X\times\U$ is the state and action space (see Section~\ref{sec:psss}).
    Algorithms not covered in \cite{deisenroth2013survey} have a lighter (green) background. References to the main paper for each of these algorithms is given in a table in the end of each section.
    From the left to the right, algorithms are grossly ranked in order of increasing sample reuse, but methods using a utility model in \Params and $O$ show better sample choice, resulting in competitive sample efficiency.}
\end{figure*}

Resulting from the above perspective, this survey is structured following the organization of methods depicted in \figurename~\ref{fig:orga_paper}.
The rest of the paper describes the different nodes in the trees and highlights some of their sample efficiency factors.
\new{A table giving a quick reference to the main paper for each algorithm is given in the end of each section.}

In Sections~\ref{sec:mfo} to \ref{sec:psss}, we present Solutions~\ref{sol1} to \ref{sol3} above in more detail, showing how the corresponding methods are implemented in various \pse algorithms.
We do not cover Solution~\ref{sol4} and refer readers to \cite{chatzilygeroudis2017black} for a recent presentation of these {\em model-based policy search} methods.
Then, in Section~\ref{sec:choices}, we discuss the different elementary design choices that matter in terms of sample efficiency.
Finally, Section~\ref{sec:conclu} summarizes the paper and provides some perspectives about current trends in the domain.


\section{Policy search without a utility model}
\label{sec:mfo}

When the function to be optimized is available but has no favorable property, the standard optimization method known as Gradient Descent consists in iteratively following the gradient of this function towards a local optimum. When the same function is only known through a model built by regression from a batch of samples, one can also do the same, but computing the gradient requires evaluations over the whole batch, which can be computationally expensive. An alternative known as Stochastic Gradient Descent (SGD) circumvents this difficulty by taking a small subset of the batch at each iteration \citep{bottou2012stochastic}. Before starting to present these methods in Section~\ref{sec:theta}, we first investigate a family of methods which perform policy search without learning a model of the utility function at all.
They do so by sampling the policy parameter space \Params and moving towards policy parameters \params of higher utility $J(\params)$.

\subsection{\new{Truly} random search}

At one extreme, the simplest black-box optimization (BBO) method randomly searches \Params until it stumbles on a good enough utility. \new{We call this method ``Truly random search'' as the name ``random search'' is used in the optimization community to refer to gradient-free methods~\cite{rastrigin1963convergence}.}
Its distinguishing feature is in its sample choice strategy: the utility of the previous \params has no impact on the choice of the next \params.

Quite obviously, this sample choice strategy is not efficient, but it requires no assumption at all on the function to be optimized.
Therefore, it is an option when this function does not show any {\em regularity} that can be exploited.
All other methods rely on the implicit assumption that $J(\params)$ presents some smoothness around optima $\params^*$,
which is a first step towards using a gradient.

\rephrased{So globally, this method provides a proof of concept that an agent can obtain a better utility without estimating any gradient at all.}
\new{Recently, other forms of gradient-free methods called {\em random search} though they are not truly random have been shown to be competitive with \drl \citep{mania2018simple}.}

\begin{figure*}[hbt!]
  \begin{center}
    \includegraphics[width=0.9\textwidth]{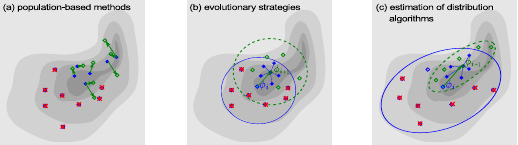}
    \caption{One iteration of evolutionary methods. (a) Population-based methods (b) Evolutionary Strategies (c) \edas. Blue: current generation and sampling domain. Full blue dots: samples with a good evaluation. Dots with a red cross: samples with a poor evaluation. Green: new generation and sampling domain, empty dots are not evaluated yet. Red dots: optimum guess. In population-based methods, the next generation are offspring from several elite individuals of the previous generation. In ES, it is obtained from an optimum guess and sampling from fixed Gaussian noise. In \edas, Gaussian noise is tuned using Covariance Matrix Adaptation.\label{fig:evo}}
  \end{center}
\end{figure*}

The next three families of methods, population-based optimization, evolutionary strategies and estimation of distribution algorithms, are all variants of {\em evolutionary} methods.
An overview\footnote{A blog with dynamical visualizations and more technical details can be found at \url{http://blog.otoro.net/2017/10/29/visual-evolution-strategies/}.} of these methods is depicted in \figurename~\ref{fig:evo}.

\subsection{Population-based optimization}
\label{sec:pop}

Population-based BBO methods manage a limited population of individuals, and generate new individuals randomly in the vicinity of the previous {\em elite} individuals.
\new{There are several families of population-based optimization methods, the most famous being Genetic Algorithms (GAs) \citep{goldberg1989genetic}, Genetic Programming (GP) \citep{koza1992genetic}, and the more advanced NEAT framework \citep{stanley2002efficient}.}
In these methods, the parameter \params corresponding to an individual is often called its {\em genotype} and the corresponding utility is called its {\em fitness}, see \cite{back1996evolutionary} for further reading.
\new{These methods have already been combined with neural networks giving rise to {\em neuroevolution} \citep{floreano2008neuroevolution} but}, up to recently, these methods were mostly applied to small to moderate size policy representations.
However, the availability of modern computational resources have made it possible to apply them to large and deep neural network representations, defining the emerging domain of {\em deep neuroevolution} \citep{such2017deep}.
Among other things, it was shown that, given large enough computational resources, \new{methods as simple as GAs} can offer a competitive alternative to \drl methods presented in Section~\ref{sec:psss}, mostly due to their excellent parallelization capabilities \citep{such2017deep,conti2017improving}.

\subsection{Evolutionary strategies}
\label{sec:es}

Evolutionary strategies (ES) can be seen as specific population-based optimization methods where only one individual is retained from one generation to the next. More specifically, an optimum guess is computed from the previous samples, then the next samples are obtained by adding Gaussian noise to the current optimum guess.

Moving from one optimum guess to the next implements a form of policy improvement similar to SGD, but where the gradient is approximated by averaging over samples instead of being analytically computed. Hence this method is more flexible but, since gradient approximation uses a random exploration component, it is less data efficient.
However, data efficiency can be improved by reusing samples between one generation and the next when their sampling domain overlaps, a method called {\em importance mixing} \citep{sun2009efficient}.
\removed{Surprisingly, importance mixing has not been incorporated into recent algorithms of the same family, maybe because the primary focus of this line of research is not sample efficiency.}
\new{An improved version of importance mixing was recently proposed in \cite{pourchot2018importance}, showing a large impact on sample efficiency, but not large enough to compete with \drl methods on this aspect. Further results about importance mixing can be found in \citep{pourchot2018cem}, showing that more investigations are necessary to better understand in which context this mechanism can be most useful.}

The correlation between the direction of the gradient given by SGD and the same direction for ES depends on the evolutionary algorithm.
Interestingly, good ES performance can be obtained even when the correlation is not high, though this result still needs to be confirmed in the case of \pse \citep{zhang2017relationship}.

A specific ES implementation of deep neuroevolution where constant Gaussian noise is used at each generation was shown to compete with \drl methods on standard benchmarks \citep{salimans2017evolution}. This simple implementation generated an insightful comparison with methods based on SGD depending on various gradient landscapes, showing under which conditions ES can find better optima than SGD \cite{lehman2017more}.

Finally, instead of approximating the vanilla gradient of utility, \nes \citep{wierstra2008natural} and \xnes \citep{glasmachers2010exponential} approximate its natural gradient \citep{akimoto2010bidirectional}, but for doing so they have to compute the inverse of the Fisher Information Matrix, which is prohibitively expensive in large dimensions \citep{grondman2012survey}.
We refer the reader to \cite{pierrot2018first} for a detailed presentation of {\em natural gradient} and other advanced gradient descent concepts.

\subsection{Estimation of Distribution Algorithms}
\label{sec:edas}

The standard perspective about \edas is that they are a specific family of ES using a covariance matrix $\covar$ \cite{larranaga2001estimation}.
This covariance matrix defines a multivariate Gaussian function over \Params, hence its size is $|\params|^2$.
Samples at the next iteration are drawn with a probability proportional to this Gaussian function.
Along iterations, the ellipsoid defined by $\covar$ is progressively adjusted to the top part of the hill corresponding to the local optimum $\params^*$.

The role of $\covar$ is to control exploration. The exploration policy can be characterized as {\em uncorrelated} when it only updates the diagonal of $\covar$ and {\em correlated} when it updates the full $\covar$ \citep{deisenroth2013survey}. The latter is more efficient in small parameter spaces but computationally more demanding and potentially inaccurate in larger spaces as more samples are required. In particular, it cannot be applied in the deep neuroevolution context where the order of magnitude of the size of \params is between thousands and millions.

Various instances of \edas, such as \cem, \cmaes, \picma, are covered in \cite{stulp12icml,stulp2012policy,stulp13paladyn}. Among them, the \cmaes algorithm is also shown to approximate the natural gradient \citep{arnold11informationgeometric}.
By contrast, the \PIBB algorithm, also described in \cite{stulp13paladyn}, is a simplification of \picma where covariance matrix adaptation has been removed.
Thus it should be considered an instance of the former ES category.

\subsection{Finite difference methods}

In {\em finite difference} methods, the gradient of utility with respect to \params is estimated as the first order approximation of the Taylor expansion of the utility function.
This estimation is performed by applying local perturbations to the current input.
Thus these methods are derivative-free and we classify them as using no model, even if they are based on a local linear approximation of the gradient.

In finite difference methods, gradient estimation can be cast as a standard regression problem, but perturbations along each dimension of \Params can be treated separately, which results in a very simple algorithm \cite{riedmiller08evaluation}. The counterpart of this simplicity is that it suffers from a lot of variance, so in practice the methods are limited to deterministic policies.

\subsection{Reference to the main algorithms}


\begin{table}[!htbp]
 \begin{center}  
   \begin{tabular}{ll}
      \hline
      Algorithm &Main paper\\
     \hline\hline
     \cmaes & \cite{hansen01completely}\\        
     \cem & \cite{rubinstein04crossentropy}\\    
     finite diff. & \cite{riedmiller08evaluation}\\
     \nes & \cite{wierstra2008natural}\\
     \hline
     \xnes & \cite{glasmachers2010exponential}\\ 
     \PIBB & \cite{stulp2012policy}\\            
     \picma & \cite{stulp12icml}\\               
     OpenAI-ES & \cite{salimans2017evolution}\\
     Random Search & \cite{mania2018simple}\\
     \hline                                                                                     
\end{tabular}
    \caption{Main gradient-free algorithms. Above the line, they were studied in \cite{deisenroth2013survey}, below they were not.
      \label{tab:gf}}
 \end{center}
\end{table}

\subsection{Sample efficiency analysis}

In all gradient-free methods, sampling a vector of policy parameters \params provides an exact information about its utility $J(\params)$.
However, the $J$ function can be stochastic, in which case one value of $J(\params)$ only contains partial information about the value of that \params.
Anyways, sample reuse can be implemented by storing an archive of the already sampled pairs $<\params,J(\params)>$.
Each time an algorithm needs the utility $J(\params)$ of a sample \params, if this utility is already available in the archive, it can use it instead of sampling again.
In the deterministic case, using the stored value is enough. In the stochastic case, the archive may provide a distribution over values $J(\params)$, and the algorithm may either draw a value from this distribution or sample again, depending on accuracy requirements. 

\begin{mymes}{mes:mf}
  Policy search without a utility model is generally less data efficient than Stochastic Gradient Descent (SGD).
  Though sample reuse is technically possible without a utility model, in practice it is seldom used.
  Despite their lower sample efficiency in comparison to SGD, some of these methods are highly parallelizable and offer a viable alternative to \drl provided enough computational resources.
\end{mymes}


\section{Policy search with a model \new{of utility} in the space of policy parameters}
\label{sec:theta}

\new{As outlined in the introduction, the utility of a vector of policy parameters is only available by observing the corresponding behavior.
Although no model that relates policy parameters to utilities is given, one may approximate the utility function in \Params from these observations, by collecting samples consisting of (policy parameters, utility) pairs and using regression to infer a model of the corresponding function (see e.g. \cite{stulp15NN}). Such a model could either be deterministic, giving one utility per policy parameters vector, or stochastic, giving a distribution over utility values.}

\rephrased{
  Once such a model is learned, one could perform gradient descent on this model. These steps could be performed sequentially (first model learning and then gradient descent) or incrementally (improving the model and performing gradient descent after each new utility observation). In the latter case, the model is necessarily {\em persistent}: it evolves from iteration to iteration given new information, in contrast with the sequential case where it could be {\em transient}, that is recomputed from scratch at each iteration.
  }

\subsection{Bayesian Optimization}

Though the above approach seems appealing, we are not aware of any algorithm performing what is described above \new{in the deterministic case}. A good reason for this is that utility functions are generally stochastic in \Params. Thus, algorithms which learn a model $J(\params)$ have to learn a distribution over such models.
This is exactly what Bayesian optimization (BO) does. The {\em distribution} over models is updated through Bayesian inference. It is initialized with a {\em prior}, and each new sample, considered as some new {\em evidence}, helps adjusting the model distribution towards a peak at the true value, whilst keeping track of the variance over models.
By estimating the uncertainty over the distribution of models, BO methods are endowed with active learning capabilities, dramatically improving their sample efficiency at the cost of a worse scalability.

A BO algorithm comes with a {\em covariance function} that determines how the information provided by a new sample influences the model distribution around this sample.
It also comes with an {\em acquisition function} used to choose the next sample given the current model distribution. A good acquisition function should take into account the value and the uncertainty of the model over the sampled space.
\removed{Furthermore, finding the optimum over the acquisition function should be computationally cheap, so as to control the computational cost of choosing the next sample.}

By quickly reducing uncertainty, BO implements a form of active learning.
As a consequence, it is very sample efficient when the parameter space is small enough, and it searches for the global optimum, rather than a local one.
However, given the necessity to optimize globally over the acquisition function, it scales poorly in the size of the parameter space.
For more details, see \cite{brochu2010tutorial}.

The \rock algorithm is an instance of BO that searches for a local optimum instead of a global one \cite{hwangbo2014rock}.
It uses \cmaes to find the optimum over the model function.
By doing so, it performs natural rather than \vgo, but it does not use the available model of the utility function, though this could improve sample efficiency.

Bayesian optimization algorithms generally use Gaussian kernels to efficiently represent the distribution over models. However, some authors have started to note that, in the specific context of \pse, BO was not using all the information available in elementary steps of the agent. This led to the investigation of more appropriate data-driven kernels based on the Kullbak-Leibler divergence between \roll density generated by two policies \citep{wilson2014using}.

Using BO in the context of policy search is an emerging domain \citep{lizotte2007automatic,calandra2014bayesian,metzen2015bayesian,martinez2017policy}. Furthermore, recent attempts to combine BO with reinforcement learning approaches, giving rise to the Bayesian Optimization Reinforcement Learning (BORL) framework, are described in Section~\ref{sec:psss}.

\newpage 
\subsection{Reference to the main algorithms}


\begin{table}[!ht]
 \begin{center}  
   \begin{tabular}{ll}
      \hline
      Algorithm &Main paper\\
     \hline\hline
     Bayes. Opt. & \cite{pelikan1999boa}\\
     \rock & \cite{hwangbo2014rock}\\
     \hline                                                                                     
\end{tabular}
    \caption{Main Bayesian Optimization algorithms
      \label{tab:bo}}
 \end{center}
\end{table}

\subsection{Sample efficiency analysis}

\new{Learning a model of the utility function in \Params should be more sample efficient than trying to optimize without a model, as the gradient with respect to the model can be used to accelerate parameter improvement.
  However, learning a deterministic model is not enough for most cases, as the true utility function is generally stochastic in \Params, and learning a stochastic model comes with an additional computational cost which impacts the scalability of the approach.}

\begin{mymes}{mes:BO2}
  Bayesian Optimization is BBO managing a distribution over models in the policy parameter space.
  Its sample efficiency benefits from active choice of samples. But as it performs global search, it does not scale well to large policy parameter spaces.
  Thus, it is difficult to apply to deep neural network representations.
\end{mymes}

\section{Directed exploration methods}
\label{sec:geps}

Directed exploration methods are particularly useful in tasks with {\it sparse rewards}, i.e. where large parts of the search space have the same utility signal.  
These methods have two main features. First, instead of searching directly in the policy parameter space \Params, they search in a smaller {\em outcome space} $O$ (also called {\em descriptor space} or {\em behavioral space}) and learn an invertible mapping between \Params and $O$. Second, they all optimize a task-independent criterion called novelty or diversity which is used to efficiently cover the outcome space.

\begin{figure}[hbt!]
  \begin{center}
    \includegraphics[width=0.99\columnwidth]{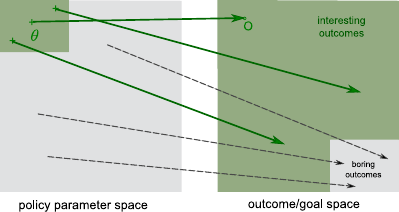}
    \caption{A standard mapping between a policy parameter space \Params and an outcome space $O$. Most often, many policy parameters result in the same outcome (for instance, in the case of a robotic arm which must move a ball around, if the policy defines arm movements and the outcome space is defined as ball positions, most policy parameters will results into a static ball). In that case, sampling directly in \Params works poorly: you have to sample in such a way to efficiently cover~$O$.
      \label{fig:gep_expli}}
  \end{center}
\end{figure}

The outcome itself corresponds to properties of the observed behavior.
The general intuition is that if the outcome space is properly covered by known policy parametrizations, and if utility can be easily related to outcomes, then it should be easy to find policy parameters with a high utility, even when the utility function is null for most policy parameters. \figurename~\ref{fig:gep_expli} visualizes why it is generally more efficient to perform the search for novel solutions in a dedicated outcome space and learning a mapping from \Params to $O$ than performing this search directly in \Params \citep{baranes2014effects}.

So, for the method to work, the outcome space has to be defined in such a way that determining the utility corresponding to an outcome is straightforward. Generally, the outcome space is defined by an external user to meet this requirement. Nevertheless, using representation learning methods to let the agent autonomously define its own outcome space is an emerging topic of interest \citep{pere2018unsupervised,laversanne2018curiosity}.

Directed exploration methods can be split into {\em novelty search} (NS) \citep{lehman2011abandoning}, {\em quality-diversity} (QD) \citep{pugh2015confronting} and {\em goal exploration processes} (\geps) \citep{baranes2010intrinsically,forestier2016overlapping,forestier2017intrinsically}. The first two derive from evolutionary methods, whereas \geps come from the developmental learning and intrinsic motivation literature.

An important distinction between them is that NS and \geps are designed to optimize diversity only\footnote{Hence the dotted line in \figurename~\ref{fig:orga_paper}.}, thus they do not use the utility function at all, whereas QD methods rely on multi-objective optimization methods to optimize diversity and utility at the same time.

The NS approach arose from the realization that optimizing utility as a single objective is not the only option \citep{doncieux2014beyond}. In particular, in the case of sparse or deceptive reward problems, it was shown that seeking novelty or diversity is an efficient strategy to obtain high utility solutions, even without explicitly optimizing this utility \citep{lehman2011abandoning}. The \gep approach was more inspired by thoughts on intrinsic motivations, where the goal was to have an agent achieve its own goal without an external utility signal \citep{forestier2017intrinsically}.
However, researchers in evolutionary methods also realized that diversity and utility can be optimized jointly \citep{cuccu2011novelty}, giving rise to more advanced NS and QD algorithms \citep{pugh2015confronting,cully2017quality}.

All these methods share a lot of similarities. They all start with a random search stage and, when they evaluate a policy parameter vector \params resulting in a point $o$ in the outcome space $O$, they store the corresponding $<\params,o>$ pair in an archive. Because they use this archive for policy improvement, they all implement a form of {\em lazy learning}, endowing them with interesting sample efficiency properties \citep{aha1997editorial}. The archive itself can be seen as a stochastic model of the function relating \Params to $O$, as made particularly obvious in the {\em MAP-Elites} algorithm \citep{cully2015robots}.

In more details, the main differences between these methods lie in the way they cover the outcome space $O$. NS and QD methods perform undirected variations to the elite \params vectors present in the archive.
\new{More precisely, in NS, the resulting solution is just added to the archive, whereas in QD the new solution replaces a previous one if it outperforms it both in terms of diversity and utility.}
By contrast, \geps choose a desired outcome $o^*$ and modify a copy of the \params leading to the closest outcome in the archive. The choice of a desired outcome $o^*$ can be performed randomly or using curriculum learning or learning progress concepts \citep{baranes2013active,forestier2017intrinsically}. Similarly, the modification of \params can be performed using undirected Gaussian noise or in more advanced ways. For instance, some \gep methods build a local linear model of the mapping from \Params to $O$ to efficiently invert it, so as to find the $\params^*$ corresponding to the desired outcome $o^*$ \citep{baranes2013active}.

Thus directed exploration methods all learn a stochastic and invertible mapping between \Params and $O$. When they also learn a model of $J(o)$, this model is stochastic with respect to \Params, which makes them similar to BO methods. In that case, the outcome space is an intermediate space between \Params and utilities: policy parameters are first projected into the outcome space, and then a model of the utility function in this outcome space can be learned.

Learning a model of the utility in $O$ shares some similarities with learning a critic in the state action space $\X\times\U$, as presented in Section~\ref{sec:psss}.
From this perspective, these methods can be seen as providing an intermediary family between evolutionary, BO and reinforcement learning methods.
However, we shall see soon that learning a critic in the state action space $\X\times\U$ benefits from additional properties related to temporal difference learning, which limits the use of the above unifying perspective.

\subsection{Reference to the main algorithms}

\begin{table}[!htbp]
 \begin{center}  
   \begin{tabular}{ll}
      \hline
      Algorithm &Main paper\\
     \hline\hline
     Novelty Search & \cite{lehman2011abandoning}\\
     Quality-Diversity & \cite{pugh2015confronting}\\
     Goal Exploration & \cite{forestier2017intrinsically}\\
     \hline                                                                                     
\end{tabular}
    \caption{Main directed exploration algorithms
      \label{tab:explo}}
 \end{center}
\end{table}

\subsection{Sample efficiency analysis}

The defining characteristic of all directed exploration methods is their capability to widely cover the outcome space. This provides efficient exploration, which in turn critically improves sample efficiency when combined with more standard evolutionary methods mentioned in Section~\ref{sec:mfo} \citep{conti2017improving} or \drl method mentioned in Section~\ref{sec:psss} \cite{colas2018gep}.

Even though our article focuses on single-task learning, it is worth mentioning that direct exploration methods may very much improve sample efficiency in multi-task learning scenarios. This is because such methods aim at covering the (interesting) outcome space, and can thus more easily adapt when facing multiple tasks, and thus potentially multiple outcomes.

\begin{mymes}{mes:exploration}
  Looking for diversity only in a user defined outcome space is an efficient way to perform exploration, and can help solve sparse or deceptive reward problems, where more standard exploration would fail. Directed exploration methods are thus useful complements to other methods covered in this survey.
\end{mymes}

\section{Policy search with a critic}
\label{sec:psss}

The previous two sections have presented methods which learn mappings from policy parameter space \Params to utilities or outcomes. We now cover methods which learn a model of utility in the state-action space $\X\times\U$.

An important component in the RL formalization, the utility $U(\vx,\vu)$ corresponds to the return the agent may expect from performing action $\vu$ when it is in state $\vx$ and then following either its current policy $\pol_\params$ or the optimal policy $\pol^*$. This quantity may also depend on a discount factor $\gamma$ and a noise parameter~$\beta$.

The true utility $U(\vx,\vu)$ can be approximated with a model $\hatu_\eta(\vx,\vu)$ with parameters $\eta$.  Such a model is called a {\em critic}. A key feature is that the critic can be learned from samples corresponding to {\em single steps} in the rollouts of the agent, either with temporal differencing or Monte Carlo methods.
Methods that approximate $U$ by $\hatu_\eta$, and determine the policy parameters \params by descending the gradient of \params with respect to $\hatu_\eta$ are called {\em actor-critic} methods, the policy $\pol_\params$ being the actor~\citep{peters2008reinforcement,deisenroth2013survey}.

This actor-critic approach can be applied to stochastic and deterministic policies \cite{silver2014deterministic}.
The space of deterministic policies being smaller than the space of stochastic policies, the latter can be advantageous because searching the former is faster than searching the latter. However, a stochastic policy might be more appropriate when Markov property does not hold \cite{williams1998experimental,sigaud2010markov} or in adversarial contexts \cite{wang2016sample}.

\subsection{Exploration in parameter or state-action space}
\label{sec:explo_sb}

As mentioned in Section~\ref{sec:theta}, learning a model of the utility in the space \Params is a regression problem that is performed by sampling and exploring directly in \Params.
In contrast, $\X\times\U$ cannot be sampled directly, as one does not know in advance which policy parameters will result in visiting which states and performing which action.
Exploration is therefore performed in either by adding noise to the \params (policy parameter perturbation), or adding noise to the actions the policy outputs (action perturbation). In the latter case, exploration is generally undirected and adds Gaussian noise or correlated Ornstein-Ulhenbeck noise to the actions taken by the policy.
Policy parameter perturbation is done in \pepg, \power and \pis, and more recently to \ddpg \citep{fortunato2017noisy,plappert2017parameter}, whereas action perturbation in the other algorithms presented in this paper.


\newcommand{\transient}{transient\xspace}
\newcommand{\persistent}{persistent\xspace}

\new{
All actor-critic algorithms iterate over the following three steps: 
\begin{itemize}
  \item[A] Collect new step samples from the current policy with policy parameter perturbation or action perturbation for exploration,
  \item[B] Compute a new critic $\hatu_\eta$ based on these samples, by determining $\eta$ through a temporal difference method,
  \item[C] Update the policy parameters \params through gradient descent with respect to the critic.
\end{itemize}
}
\new{
A distinction here should be made on whether 1)~the critic is discarded after step C, and must thus be learned from scratch in step B in the next iteration, or 2)~the critic is persistent throughout the learning, and incrementally updated in step B. We discuss the differences between these two variants -- which we denote {\it \transient critic} and {\it \persistent critic} respectively -- in more detail in the next two sections.
}


\subsection{Transient Critic Algorithms}
\label{sec:iter_ac}

\removed{Methods following the iterative approach can be characterized as {\em policy iteration} methods: they learn a new critic at each iteration \cite{schulman2015trust}.
As outlined in \figurename~2 of \cite{stulp13paladyn}, algorithms from this family are similar to \edas, apart from the fact that they model the utility function in $\X\times\U$ instead of \Params.}
In methods with a transient critic, Monte Carlo sampling -- running a large set of episodes and averaging over the stochastic return -- is used to evaluate the current policy and generate a new set of step samples. Then, determining the optimal critic parameters given these samples can be cast as a batch regression problem.

Among these methods, one must distinguish between three families: {\em likelihood ratio} methods such as \reinforce \citep{williams1992} and \pepg \citep{sehnke2010parameter}, {\em natural gradient} methods such as \nac and \enac~\footnote{The critic is generally persistent in actor-critic methods, but this is not the case in \nac and \enac.} \citep{peters08natural} and {\em EM-based} methods such as \power\footnote{Interestingly, \pis can also be seen as a transient critic method, though it could in principle use a persistent oneand fall into Section~\ref{sec:onl}. This is just because using batch updates make it more stable \citep{deisenroth2013survey}.} \citep{kober2009learning} and the variants of \reps \citep{peters2010relative}. All the corresponding algorithms are well described in \cite{deisenroth2013survey}.

Although they derive from a different mathematical framework, likelihood ratio methods and EM-based methods are similar: they both use unbiased estimation of the gradient through Monte Carlo sampling and they are both mathematically designed so that the most rewarding \rolls get the highest probability.

The \trpo \cite{schulman2015trust} algorithm also follows an iterative approach and can use a deep neural network representation, thus it can be classified as a \drl method.
Among other things, it uses a bound on the Kullback-Leibler divergence between policies at successive iterations to ensure safe and efficient exploration.
Finally, the Guided Policy Search (\gps) algorithm \cite{levine2013guided,montgomery2016guided} is another transient critic \drl method inspired from \reinforce, but adding guiding \rolls obtained from simpler policies.

\subsection{Persistent Critic Algorithms}
\label{sec:onl}

In contrast with transient critic algorithms, persistent critic algorithms incrementally update the critic during training. \removed{All such algorithms use an actor-critic architecture.}
\new{Most such algorithms use an actor-critic architecture, with the notable exception of \naf \citep{gu2016continuous}, which does not have an explicit representation of the actor.}
To our knowledge, before the emergence of \drl algorithms described below, the four \inac algorithms were the only representative of this family \citep{bhatnagar07_NIPS}.

The way to compute the critic incrementally can be named a {\em temporal difference} (TD) method, also named a {\em bootstrap} method \citep{sutton88}.
They compute at each step a {\em \td error} or {\em reward prediction error} (RPE) between the immediate reward predicted by the current values of the critic and the actual reward received by the agent. This RPE can then be used as a loss that the critic should minimize over iterations \citep{sutton98}.

\subsection{Key properties of Persistent Critic Algorithms}

\new{Most mechanisms that made deep actor-critic algorithms possible where first introduced in \dqn \citep{mnih2015human}. Though \dqn is a discrete action algorithm which is outside the scope of this survey, we briefly review its important concepts and mechanisms before listing the main algorithms in the family of continuous action \drl methods.
  }

\subsubsection{Accuracy and scalability: deep neural networks}

By using deep neural networks as approximation functions and making profit of large computational capabilities of modern clusters of computers, all \drl algorithms are capable of addressing much larger problems than before, and to approximate gradients with unprecedented accuracy, which makes them more stable than the previous linear architectures of \nac and \power, hence amenable to incremental updates of a persistent critic rather than recomputing a transient one.

\subsubsection{Stability: the target critic}

\Drl methods introduced a {\em target critic} as a way to improve stability.
Standard regression is the process of fitting samples to a model so as to approximate an unknown stationary function \citep{stulp15NN}.
Estimating a critic through temporal difference methods is similar to regression, but the target function is not stationary: it is itself a function of the estimated critic, thus it is modified each time the critic is modified. This can result in divergence of the critic when the target function and the estimated critic are racing after each other \citep{baird94}.
To mitigate this instability, one should keep the target function stationary during several updates and reset it periodically to a new function corresponding to the current critic estimate, switching from a regression problem to another. This idea was first introduced in \dqn \citep{mnih2015human} and then modified from periodic updates to smooth variations in \ddpg \citep{lillicrap2015continuous}.

\subsubsection{Sample reuse: the replay buffer}

Since they are based on {\em value propagation}, TD methods can give rise to more sample reuse than standard regression methods, provided that these samples are saved into a {\em replay buffer}.
Using a replay buffer is at the heart of the emergence of modern actor-critic approaches in \drl.
Actually, learning from the samples in the order in which they are collected is detrimental to learning performance and stability
because these samples are not independent and identically distributed (i.i.d.).
Stability is improved by drawing the samples randomly from the replay buffer and sample efficiency
can be further improved by better choosing the samples, using {\em prioritized experience replay} \citep{schaul2015prioritized}.


\subsubsection{Adaptive step sizes and return lengths}

Modern SGD methods provided by most machine learning libraries now incorporate adaptive step sizes, removing a difficulty with previous actor-critic algorithms such as \enac. Another important ingredient for the success of some recent methods in the use of {\em n-step return}, which consists in performing temporal difference updates over several time steps, resulting in the possibility to control the bias-variance trade-off (see Section~\ref{sec:bias_var}).

\subsection{Overview of \drl algorithms}

All these favorable properties are common traits of several incremental \drl algorithms: \ddpg \citep{lillicrap2015continuous}, \naf \citep{gu2016continuous}, \ppo \cite{schulman2017proximal}, \acktr \citep{wu2017scalable}, \sac \citep{haarnoja2018soft}, \tddd \citep{fujimoto2018addressing} and \dfpg \citep{ddddpg}. As depicted in \figurename~\ref{fig:orga_paper}, the last one, \dfpg, is an instance of Bayesian Optimization Reinforcement Learning (BORL) algorithms which derive from BO but belong to the step-based category of methods described in Section~\ref{sec:psss}. These algorithms result from an effort to incorporate Bayesian computations into the \drl framework, and correspond to a very active trend in the field.
Most of these works address discrete actions \citep{azizzadenesheli2018efficient,tang2017variational}, but \dfpg is an exception that derives from adopting a distributional perspective on policy gradient computation, resulting in more accurate estimates on the gradient and better sample efficiency \citep{bellemare2017distributional}.

Finally, a few algorithms such as \acer \citep{wang2016sample}, \qprop \citep{gu2016q} and \pgql \citep{o2016pgq} combine properties of transient and persistent critic methods, and are captured into the more general framework of Interpolated Policy Gradient (\ipg) \citep{gu2017interpolated}.
For a more detailed description of all these algorithms, we refer the reader to the corresponding papers and to a recent survey \citep{arulkumaran2017brief}.


\subsection{Reference to the main algorithms}

\begin{table}[!htbp]
 \begin{center}  
   \begin{tabular}{ll}
      \hline
      Algorithm &Main paper\\
     \hline\hline
     \reinforce & \cite{williams1992}\\       
     \gpomdp & \cite{baxter2001infinite}\\
     \nac & \cite{peters08natural}\\          
     \enac & \cite{peters08natural}\\         
     \power & \cite{kober2009learning}\\      
     \pis & \cite{theodorou10generalized}\\
     \reps & \cite{peters2010relative}\\
     \pepg & \cite{sehnke2010parameter}\\     
     \vips & \cite{neumann2011variational}\\    

     \hline                                                                                     
 
     \inac & \cite{bhatnagar07_NIPS}\\        
     \gps & \cite{levine2013guided}\\         
     \trpo & \cite{schulman2015trust}\\       
     \ddpg & \cite{lillicrap2015continuous}\\ 
     \aaac & \cite{mnih2016asynchronous}\\    
     \naf & \cite{gu2016continuous}\\         
     \acer & \cite{wang2016sample}\\          
     \qprop & \cite{gu2016q}\\                
     \pgql & \cite{o2016pgq}\\
     \ppo & \cite{schulman2017proximal}\\
     \acktr &\citep{wu2017scalable}\\
     \sac & \citep{haarnoja2018soft}\\
     \tddd & \cite{fujimoto2018addressing}\\
     \dfpg & \cite{ddddpg}\\
     \hline                                                                                     
\end{tabular}
    \caption{Main reinforcement learning algorithms. Algorithms below the line have not yet been covered in \cite{deisenroth2013survey}.
      \label{tab:rl}}
 \end{center}
\end{table}

\subsection{Sample efficiency analysis}

\begin{mymes}{mes:replay}
  Being step-based, \drl methods are able to use more information from rollouts than episode-based methods.
  Furthermore, using a replay buffer leads to further sample reuse. 
\end{mymes}

\section{Discussion}
\label{sec:choices}

In the previous sections we have presented methods which: (1) do {\it not} build a utility model; (2) learn a utility model: (2a) in the policy parameter space \Params, (2b) in an arbitrary outcome space $O$, (2c) in the state-action space $\X\times\U$.
In this section, we come back to the sample efficiency properties of these different methods.
We do so by descending the tree of design choices depicted in \figurename~\ref{fig:orga_paper}.

\subsection{Building a model or not}

We have outlined that \pse methods which build a model of the utility function are generally more sample efficient than methods which do not.
However, the reliance of the latter to SGD can make them less robust to local optima \citep{lehman2017more} and it has been shown recently
that methods which do not build a model of utility are still competitive in terms of final performance, due to their higher parallelization capability and distinguishing
properties with respect to various gradient landscapes \cite{salimans2017evolution,such2017deep,zhang2017relationship}.

\subsection{Building a utility function model in the policy parameter space versus the state-action space.}
\label{sec:spaces}

Several elements speak in favor of the higher sample efficiency of learning a critic in the state-action space $\X\times\U$.
First, it can give rise to more sample reuse than learning a model of the utility function in \Params.
Second, learning from each step separately makes a better use of the information available from a \roll than learning from global episodes.

Furthermore, $\X\times\U$ may naturally exhibit a hierarchical structure -- especially the state --  which is not so obvious for \Params. As a consequence, methods that model utility in $\X\times\U$ may benefit from learning intermediate representations at different levels in the hierarchy, thus reducing the dimensionality of the \pse problem.
Learning such intermediate and more compact representations is the focus of hierarchical reinforcement learning, a domain which has also been impacted by the emergence of \drl \citep{kulkarni2016hierarchical,bacon2017option}. Hierarchical reinforcement learning can also be performed \ofl, which corresponds to the perspective of the DREAM project\footnote{\url{http://www.robotsthatdream.eu/}}, illustrated for instance in \citep{zimmer2017bootstrapping}.

Finally, an important factor of sample efficiency is the size and structure of $\X\times\U$ with respect to \Params.
In both respects, the emergence of \drl methods using large neural networks as policy representation has changed the perspective.
First, in \drl, the size of \Params can become larger than that of $\X\times\U$, which speaks in favor of learning a critic.
Second, deep neural networks seem to generally induce a smooth structure between \Params and $\X\times\U$, which facilitates learning.
Finally, a utility function modeled in a larger space may suffer from fewer local minima, as more directions remain for improving the gradient \cite{kawaguchi2016deep}.

The above conclusions might be mitigated by considering exploration. Indeed, in several surveys about \pse for robotics, policy parameter perturbation methods are considered superior to action perturbation methods \citep{stulp13paladyn,deisenroth2013survey}. This analysis is backed-up with several mathematical arguments, but it might be true mostly when the space \Params is smaller than the space $\X\times\U$. Until recently, all \drl methods were using action perturbation.
But \drl algorithms using policy parameter perturbation have recently been published, showing again that one can model the utility function in $\X\times\U$ while performing exploration in \Params \citep{fortunato2017noisy,plappert2017parameter}.
Exploration is currently one of the hottest topics in \drl and directed exploration methods presented in Section~\ref{sec:geps} may play a key role in this story, despite the lower data efficiency of their policy improvement mechanisms \citep{conti2017improving,colas2018gep}.

\begin{mymes}{mes:spaces}
  There are more arguments for learning a utility model in $\X\times\U$ than in \Params, but this ultimately depends on the size of these spaces and the structure
  of their relationship.
\end{mymes}

\subsection{Transient versus persistent critic}
\label{sec:iter_incr}

At first glance, having a persistent critic may seem superior to having a transient one, for three reasons. First, by avoiding to compute the critic again at each iteration,
it is computationally more efficient. Second, immediate updates favor data efficiency because the policy is improved as soon as possible, which in turn helps generating better samples. Third, being based on bootstrap methods, they give rise to more sample reuse. However, these statements must be differentiated, as two factors (described below) must be taken into account.

\subsubsection{Trading bias against variance}
\label{sec:bias_var}

Estimating the utility of a policy in $\X\times\U$ is subject to a bias-variance compromise \citep{kearns2000bias}.
On the one hand, estimating the utility of a given policy through Monte Carlo sampling -- as is generally done in transient critic approaches -- is subject to a variance which grows with the length of the episodes.
On the other hand, incrementally updating a persistent critic reduces variance, but may suffer from bias, resulting in potential sub-optimality, or even divergence.
Instead of performing bootstrap updates of a critic over one step, one can do so over $N$ steps. The larger $N$, the closer to Monte Carlo estimation, thus tuning $N$ is a way of controlling the bias-variance compromise.
For instance, while the transient critic \trpo algorithm is less sample efficient than actor-critic methods but more stable, often resulting in superior performance \citep{duan2016benchmarking}, its immediate successor, \ppo, uses $N$ steps return, resulting in a good compromise between both families \citep{schulman2017proximal}.

\subsubsection{Off-policy versus \onp updates}
\label{sec:onp_ofp}

In \onp methods such as \sarsa, the samples used to learn the critic must come from the current policy, whereas in \ofp methods such as \ql, they can come from any policy.
In most transient critic methods, the samples are discarded from one iteration to the next and these methods are generally \onp.
By contrast, persistent critic methods using a replay buffer are generally \ofp\footnote{The \aaac algorithm is an incremental actor-critic method which does not use a replay buffer, and it is classified as \onp.}.

This \onp versus  \ofp distinction is related to the bias-variance compromise.
Indeed, when learning a persistent critic incrementally, using \ofp updates is more flexible because the samples can come from any policy, but these \ofp updates introduce bias in the estimation of the critic.
As a result, \ofp methods such as \ddpg and \naf are more sample efficient because they use a replay buffer, but they are also more prone to sub-optimality and divergence.
In that respect, a key contribution of \acer and \qprop is that they provide an \ofp, sample efficient update method which strongly controls the bias, resulting in more stability \citep{gu2016q,wang2016sample,wu2017scalable,gu2017interpolated}. These aspects are currently the subject of intensive research but the resulting algorithms suffer from being more complex, with additional meta-parameters.

\begin{mymes}{mes:persistent}
  Persistent critic methods are superior to transient critic methods in many respects, but the latter are more stable because they decorrelate the problem of estimating the utility function from the problem of descending its gradient, and they suffer from less bias.
\end{mymes}

\section{Conclusion}
\label{sec:conclu}


In this paper, we have contrasted various approaches to policy search, from evolutionary methods which do not learn a model of the utility function to \drl methods which do so in the state-action space.

In \cite{stulp13paladyn}, the authors have shown that \pse applied to robotics was shifting from actor-critic methods to evolutionary methods.
Part of this shift was due to the use of open-loop \dmps \cite{ijspeert2013dynamical} as a policy representation, which favors episode-based approaches, but another part resulted from the higher stability and efficiency of evolutionary methods by that time.

The emergence of \drl methods has changed this perspective.
It should be clear from this survey that, in the context of large problems where deep neural network representations are now the standard option, \drl is generally more sample efficient than deep neuroevolution methods,
as empirically confirmed in \cite{broissia2016actor} and \cite{pourchot2018importance}.
The higher sample efficiency of \drl methods, and particularly actor-critic architectures with a persistent critic, results from several mechanisms.
They benefit from better approximation capability of non-linear critics and the incorporation of an adapted step size in SGD, they model the utility function in the state-action space, and they benefit from massive sample reuse by using a replay buffer. Using a target network has also mitigated the intrinsic instability of incrementally approximating a critic. However, it is important to acknowledge that incremental \drl methods still suffer from significant instability\footnote{As outlined at \url{https://www.alexirpan.com/2018/02/14/rl-hard.html}.}.


\subsection{Future directions}
\label{sec:future}

The field of \pse is currently the object of an intense race for increased performance, stability and sample efficiency.
We now outline what we currently consider as promising research directions.

\subsubsection{More analyses than competitions}
\label{sec:analysis}

Up to now, the main trend in the literature focuses on performance comparisons \citep{duan2016benchmarking,islam2017reproducibility,henderson2017deep,such2017deep} showing that, despite their lower sample efficiency, methods which do not build a model of utility are still a competitive alternative in terms of final performance \citep{salimans2017evolution,chrabaszcz2018back}. But stability and sample efficiency comparisons are missing and works analyzing the reasons why an algorithm performs better than another are only just emerging \citep{lehman2017more,zhang2017relationship,gangwani2017genetic}.
By drawing an overview of the whole field and revealing some important factors behind sample efficiency,
this paper was intended as a starting point towards broader and deeper analyses of the efficacy of various \pse methods.

\subsubsection{More combinations than competitions}
\label{sec:combine}

An important trend corresponds to the emergence of methods which combine algorithms from various families described above.
As already noted in Section~\ref{sec:geps}, directed exploration methods are often combined with evolutionary or \drl methods \citep{conti2017improving,colas2018gep}.
There is also an emerging trend combining evolutionary or population-based methods and \drl methods \citep{jaderberg2017population,khadka2018evolutionary,pourchot2018cem} which seem to be able to take the best of both worlds.
We believe we are just at the beginning of such combinations and that this area has a lot of potential.

\subsubsection{Beyond single policy improvement}
\label{sec:beyond}

Though we decided to keep {\em lifelong}, {\em continual} and {\em open-ended} learning outside the scope of this survey, we must mention that fast progress in policy improvement has favored an important tendency to address several tasks at the same time \citep{yang2014unified}. This subfield is extremely active at the moment, with many works in multitask learning \citep{vezhnevets2017feudal,veeriah2018many,gangwani2018policy}, Hierarchical Reinforcement Learning \citep{levy2018hierarchical,nachum2018data} and meta reinforcement learning \citep{wang2016learning}, to cite only a few.

Finally, because we focused on these elementary aspects, we have left aside the emerging topic of state representation learning \citep{jonschkowski2014state,raffin2016unsupervised,lesort2018state} or using auxiliary tasks for improving \drl \citep{shelhamer2016loss,jaderberg2016reinforcement,riedmiller2018learning}. The impact of these methods should be made clearer in the future.

\subsection{Final word}
\label{sec:final}

As we have highlighted in the article, research in \pse and \drl moves at a very high pace. Therefore, forecasting future trends, as we have done above, is risky, and even attempts to analyze the factors underlying current trends may be quickly outdated, but this also what makes this such an exciting research field.


\section*{Acknowledgments}
\noindent
Olivier Sigaud was supported by the European Commission, within the DREAM project, and has received funding from the European Union's Horizon 2020 research and innovation program under grant agreement $N^o$ 640891. 
Freek Stulp was supported by the HGF project ``Reduced Complexity Models'' and received funding from the European Commission through the project H2020 AnDy (GA no. 731540).
We thank David Filliat, Nicolas Perrin and Pierre-Yves Oudeyer for their feedback on this article.



\begin{thebibliography}{129}
\expandafter\ifx\csname natexlab\endcsname\relax\def\natexlab#1{#1}\fi
\expandafter\ifx\csname url\endcsname\relax
  \def\url#1{\texttt{#1}}\fi
\expandafter\ifx\csname urlprefix\endcsname\relax\def\urlprefix{URL }\fi

\bibitem[{Aha(1997)}]{aha1997editorial}
Aha, D.~W., 1997. Editorial. In: Lazy learning. Springer, pp. 7--10.

\bibitem[{Akimoto et~al.(2010)Akimoto, Nagata, Ono, and
  Kobayashi}]{akimoto2010bidirectional}
Akimoto, Y., Nagata, Y., Ono, I., Kobayashi, S., 2010. Bidirectional relation
  between cma evolution strategies and natural evolution strategies. In:
  International Conference on Parallel Problem Solving from Nature. Springer,
  pp. 154--163.

\bibitem[{Argall et~al.(2009)Argall, Chernova, Veloso, and Browning}]{argall09}
Argall, B.~D., Chernova, S., Veloso, M., Browning, B., 2009. A survey of robot
  learning from demonstration. Robotics and Autonomous Systems 57, 469--483.

\bibitem[{Arnold et~al.(2011)Arnold, Auger, Hansen, and
  Ollivier}]{arnold11informationgeometric}
Arnold, L., Auger, A., Hansen, N., Ollivier, Y., 2011. Information-geometric
  optimization algorithms: A unifying picture via invariance principles. Tech.
  rep., INRIA Saclay.

\bibitem[{Arulkumaran et~al.(2017)Arulkumaran, Deisenroth, Brundage, and
  Bharath}]{arulkumaran2017brief}
Arulkumaran, K., Deisenroth, M.~P., Brundage, M., Bharath, A.~A., 2017. A brief
  survey of deep reinforcement learning. arXiv preprint arXiv:1708.05866.

\bibitem[{Azizzadenesheli et~al.(2018)Azizzadenesheli, Brunskill, and
  Anandkumar}]{azizzadenesheli2018efficient}
Azizzadenesheli, K., Brunskill, E., Anandkumar, A., 2018. Efficient exploration
  through bayesian deep q-networks. arXiv preprint arXiv:1802.04412.

\bibitem[{Back(1996)}]{back1996evolutionary}
Back, T., 1996. Evolutionary algorithms in theory and practice: evolution
  strategies, evolutionary programming, genetic algorithms. Oxford university
  press.

\bibitem[{Bacon et~al.(2017)Bacon, Harb, and Precup}]{bacon2017option}
Bacon, P.-L., Harb, J., Precup, D., 2017. The option-critic architecture. In:
  AAAI. pp. 1726--1734.

\bibitem[{Baird(1994)}]{baird94}
Baird, L.~C., 1994. Reinforcement learning in continuous time: Advantage
  updating. In: Proceedings of the International Conference on Neural Networks.
  Orlando, FL.

\bibitem[{Baranes and Oudeyer(2010)}]{baranes2010intrinsically}
Baranes, A., Oudeyer, P.-Y., 2010. Intrinsically motivated goal exploration for
  active motor learning in robots: A case study. In: {IEEE/RSJ International
  Conference on Intelligent Robots and Systems (IROS 2010)}. IEEE, Taipei,
  Taiwan, Province Of China.

\bibitem[{Baranes and Oudeyer(2013)}]{baranes2013active}
Baranes, A., Oudeyer, P.-Y., 2013. Active learning of inverse models with
  intrinsically motivated goal exploration in robots. Robotics and Autonomous
  Systems 61~(1), 49--73.

\bibitem[{Baranes et~al.(2014)Baranes, Oudeyer, and
  Gottlieb}]{baranes2014effects}
Baranes, A.~F., Oudeyer, P.-Y., Gottlieb, J., 2014. The effects of task
  difficulty, novelty and the size of the search space on intrinsically
  motivated exploration. Frontiers in neuroscience 8, 317.

\bibitem[{Barth-maron et~al.(2018)Barth-maron, Hoffman, Budden, Dabney, Horgan,
  TB, Muldal, Heess, and Lillicrap}]{ddddpg}
Barth-maron, G., Hoffman, M., Budden, D., Dabney, W., Horgan, D., TB, D.,
  Muldal, A., Heess, N., Lillicrap, T.~P., 2018. Distributional policy
  gradient. In: ICLR. pp. 1--16.

\bibitem[{Baxter and Bartlett(2001)}]{baxter2001infinite}
Baxter, J., Bartlett, P.~L., 2001. Infinite-horizon policy-gradient estimation.
  Journal of Artificial Intelligence Research 15, 319--350.

\bibitem[{Bellemare et~al.(2017)Bellemare, Dabney, and
  Munos}]{bellemare2017distributional}
Bellemare, M.~G., Dabney, W., Munos, R., 2017. A distributional perspective on
  reinforcement learning. arXiv preprint arXiv:1707.06887.

\bibitem[{Bhatnagar et~al.(2007)Bhatnagar, Sutton, Ghavamzadeh, and
  Lee}]{bhatnagar07_NIPS}
Bhatnagar, S., Sutton, R.~S., Ghavamzadeh, M., Lee, M., 2007. Incremental
  natural actor-critic algorithms. In: Advances in Neural Information
  Processing Systems. MIT Press.

\bibitem[{Bottou(2012)}]{bottou2012stochastic}
Bottou, L., 2012. Stochastic gradient descent tricks. In: Neural networks:
  Tricks of the trade. Springer, pp. 421--436.

\bibitem[{Brochu et~al.(2010)Brochu, Cora, and De~Freitas}]{brochu2010tutorial}
Brochu, E., Cora, V.~M., De~Freitas, N., 2010. A tutorial on bayesian
  optimization of expensive cost functions, with application to active user
  modeling and hierarchical reinforcement learning. arXiv preprint
  arXiv:1012.2599.

\bibitem[{Calandra et~al.(2014)Calandra, Gopalan, Seyfarth, Peters, and
  Deisenroth}]{calandra2014bayesian}
Calandra, R., Gopalan, N., Seyfarth, A., Peters, J., Deisenroth, M.~P., 2014.
  Bayesian gait optimization for bipedal locomotion. In: International
  Conference on Learning and Intelligent Optimization. Springer, pp. 274--290.

\bibitem[{Chatzilygeroudis et~al.(2017)Chatzilygeroudis, Rama, Kaushik, Goepp,
  Vassiliades, and Mouret}]{chatzilygeroudis2017black}
Chatzilygeroudis, K., Rama, R., Kaushik, R., Goepp, D., Vassiliades, V.,
  Mouret, J.-B., 2017. Black-box data-efficient policy search for robotics.
  arXiv preprint arXiv:1703.07261.

\bibitem[{Chrabaszcz et~al.(2018)Chrabaszcz, Loshchilov, and
  Hutter}]{chrabaszcz2018back}
Chrabaszcz, P., Loshchilov, I., Hutter, F., 2018. Back to basics: Benchmarking
  canonical evolution strategies for playing atari. arXiv preprint
  arXiv:1802.08842.

\bibitem[{Colas et~al.(2018)Colas, Sigaud, and Oudeyer}]{colas2018gep}
Colas, C., Sigaud, O., Oudeyer, P.-Y., 2018. {GEP-PG}: Decoupling exploration
  and exploitation in deep reinforcement learning algorithms. arXiv preprint
  arXiv:1802.05054.

\bibitem[{Conti et~al.(2017)Conti, Madhavan, Such, Lehman, Stanley, and
  Clune}]{conti2017improving}
Conti, E., Madhavan, V., Such, F.~P., Lehman, J., Stanley, K.~O., Clune, J.,
  2017. Improving exploration in evolution strategies for deep reinforcement
  learning via a population of novelty-seeking agents. arXiv preprint
  arXiv:1712.06560.

\bibitem[{Cuccu and Gomez(2011)}]{cuccu2011novelty}
Cuccu, G., Gomez, F., 2011. When novelty is not enough. In: European Conference
  on the Applications of Evolutionary Computation. Springer, pp. 234--243.

\bibitem[{Cully et~al.(2015)Cully, Clune, Tarapore, and
  Mouret}]{cully2015robots}
Cully, A., Clune, J., Tarapore, D., Mouret, J.-B., 2015. Robots that can adapt
  like animals. Nature 521~(7553), 503--507.

\bibitem[{Cully and Demiris(2017)}]{cully2017quality}
Cully, A., Demiris, Y., 2017. Quality and diversity optimization: A unifying
  modular framework. IEEE Transactions on Evolutionary Computation.

\bibitem[{de~Froissard~de Broissia and Sigaud(2016)}]{broissia2016actor}
de~Froissard~de Broissia, A., Sigaud, O., 2016. Actor-critic versus direct
  policy search: a comparison based on sample complexity. arXiv preprint
  arXiv:1606.09152.

\bibitem[{Deisenroth and Rasmussen(2011)}]{deisenroth2011pilco}
Deisenroth, M., Rasmussen, C.~E., 2011. Pilco: A model-based and data-efficient
  approach to policy search. In: Proceedings of the 28th International
  Conference on machine learning. pp. 465--472.

\bibitem[{Deisenroth et~al.(2013)Deisenroth, Neumann, Peters,
  et~al.}]{deisenroth2013survey}
Deisenroth, M.~P., Neumann, G., Peters, J., et~al., 2013. A survey on policy
  search for robotics. Foundations and Trends{\textregistered} in Robotics
  2~(1--2), 1--142.

\bibitem[{Doncieux and Mouret(2014)}]{doncieux2014beyond}
Doncieux, S., Mouret, J.-B., 2014. Beyond black-box optimization: a review of
  selective pressures for evolutionary robotics. Evolutionary Intelligence
  7~(2), 71--93.

\bibitem[{Duan et~al.(2016)Duan, Chen, Houthooft, Schulman, and
  Abbeel}]{duan2016benchmarking}
Duan, Y., Chen, X., Houthooft, R., Schulman, J., Abbeel, P., 2016. Benchmarking
  deep reinforcement learning for continuous control. arXiv preprint
  arXiv:1604.06778.

\bibitem[{Floreano et~al.(2008)Floreano, D{\"u}rr, and
  Mattiussi}]{floreano2008neuroevolution}
Floreano, D., D{\"u}rr, P., Mattiussi, C., 2008. Neuroevolution: from
  architectures to learning. Evolutionary Intelligence 1~(1), 47--62.

\bibitem[{Forestier et~al.(2017)Forestier, Mollard, and
  Oudeyer}]{forestier2017intrinsically}
Forestier, S., Mollard, Y., Oudeyer, P.-Y., 2017. Intrinsically motivated goal
  exploration processes with automatic curriculum learning. arXiv preprint
  arXiv:1708.02190.

\bibitem[{Forestier and Oudeyer(2016)}]{forestier2016overlapping}
Forestier, S., Oudeyer, P.-Y., 2016. Overlapping waves in tool use development:
  a curiosity-driven computational model.

\bibitem[{Fortunato et~al.(2017)Fortunato, Azar, Piot, Menick, Osband, Graves,
  Mnih, Munos, Hassabis, Pietquin, et~al.}]{fortunato2017noisy}
Fortunato, M., Azar, M.~G., Piot, B., Menick, J., Osband, I., Graves, A., Mnih,
  V., Munos, R., Hassabis, D., Pietquin, O., et~al., 2017. Noisy networks for
  exploration. arXiv preprint arXiv:1706.10295.

\bibitem[{Fujimoto et~al.(2018)Fujimoto, van Hoof, and
  Meger}]{fujimoto2018addressing}
Fujimoto, S., van Hoof, H., Meger, D., 2018. Addressing function approximation
  error in actor-critic methods. arXiv preprint arXiv:1802.09477.

\bibitem[{Gangwani and Peng(2017)}]{gangwani2017genetic}
Gangwani, T., Peng, J., 2017. Genetic policy optimization. arXiv preprint
  arXiv:1711.01012.

\bibitem[{Gangwani and Peng(2018)}]{gangwani2018policy}
Gangwani, T., Peng, J., 2018. Policy optimization by genetic distillation. In:
  ICLR 2018.

\bibitem[{Gill et~al.(1981)Gill, Murray, and Wright}]{gill1981practical}
Gill, P.~E., Murray, W., Wright, M.~H., 1981. Practical optimization. Academic
  press.

\bibitem[{Glasmachers et~al.(2010)Glasmachers, Schaul, Yi, Wierstra, and
  Schmidhuber}]{glasmachers2010exponential}
Glasmachers, T., Schaul, T., Yi, S., Wierstra, D., Schmidhuber, J., 2010.
  Exponential natural evolution strategies. In: Proceedings of the 12th annual
  conference on Genetic and evolutionary computation. ACM, pp. 393--400.

\bibitem[{Goldberg(1989)}]{goldberg1989genetic}
Goldberg, D.~E., 1989. Genetic {A}lgorithms in {S}earch, {O}ptimization, and
  {M}achine {L}earning. Addison Wesley, Reading, MA.

\bibitem[{Grondman et~al.(2012)Grondman, Busoniu, Lopes, and
  Babuska}]{grondman2012survey}
Grondman, I., Busoniu, L., Lopes, G.~A., Babuska, R., 2012. A survey of
  actor-critic reinforcement learning: Standard and natural policy gradients.
  IEEE Transactions on Systems, Man, and Cybernetics, Part C (Applications and
  Reviews) 42~(6), 1291--1307.

\bibitem[{Gu et~al.(2016{\natexlab{a}})Gu, Lillicrap, Ghahramani, Turner, and
  Levine}]{gu2016q}
Gu, S., Lillicrap, T., Ghahramani, Z., Turner, R.~E., Levine, S.,
  2016{\natexlab{a}}. Q-prop: Sample-efficient policy gradient with an
  off-policy critic. arXiv preprint arXiv:1611.02247.

\bibitem[{Gu et~al.(2017)Gu, Lillicrap, Ghahramani, Turner, Sch{\"o}lkopf, and
  Levine}]{gu2017interpolated}
Gu, S., Lillicrap, T., Ghahramani, Z., Turner, R.~E., Sch{\"o}lkopf, B.,
  Levine, S., 2017. Interpolated policy gradient: Merging on-policy and
  off-policy gradient estimation for deep reinforcement learning. arXiv
  preprint arXiv:1706.00387.

\bibitem[{Gu et~al.(2016{\natexlab{b}})Gu, Lillicrap, Sutskever, and
  Levine}]{gu2016continuous}
Gu, S., Lillicrap, T., Sutskever, I., Levine, S., 2016{\natexlab{b}}.
  Continuous deep q-learning with model-based acceleration. arXiv preprint
  arXiv:1603.00748.

\bibitem[{Haarnoja et~al.(2018)Haarnoja, Zhou, Abbeel, and
  Levine}]{haarnoja2018soft}
Haarnoja, T., Zhou, A., Abbeel, P., Levine, S., 2018. Soft actor-critic:
  Off-policy maximum entropy deep reinforcement learning with a stochastic
  actor. arXiv preprint arXiv:1801.01290.

\bibitem[{Hansen and Ostermeier(2001)}]{hansen01completely}
Hansen, N., Ostermeier, A., 2001. Completely derandomized self-adaptation in
  evolution strategies. Evolutionary Computation 9~(2), 159--195.

\bibitem[{Henderson et~al.(2017)Henderson, Islam, Bachman, Pineau, Precup, and
  Meger}]{henderson2017deep}
Henderson, P., Islam, R., Bachman, P., Pineau, J., Precup, D., Meger, D., 2017.
  Deep reinforcement learning that matters. arXiv preprint arXiv:1709.06560.

\bibitem[{Hessel et~al.(2017)Hessel, Modayil, Van~Hasselt, Schaul, Ostrovski,
  Dabney, Horgan, Piot, Azar, and Silver}]{hessel2017rainbow}
Hessel, M., Modayil, J., Van~Hasselt, H., Schaul, T., Ostrovski, G., Dabney,
  W., Horgan, D., Piot, B., Azar, M., Silver, D., 2017. Rainbow: Combining
  improvements in deep reinforcement learning. arXiv preprint arXiv:1710.02298.

\bibitem[{Hwangbo et~al.(2014)Hwangbo, Gehring, Sommer, Siegwart, and
  Buchli}]{hwangbo2014rock}
Hwangbo, J., Gehring, C., Sommer, H., Siegwart, R., Buchli, J., 2014.
  {ROCK}$^{*}$:efficient black-box optimization for policy learning. In:
  IEEE-RAS International Conference on Humanoid Robots. IEEE, pp. 535--540.

\bibitem[{Ijspeert et~al.(2013)Ijspeert, Nakanishi, Hoffmann, Pastor, and
  Schaal}]{ijspeert2013dynamical}
Ijspeert, A.~J., Nakanishi, J., Hoffmann, H., Pastor, P., Schaal, S., 2013.
  Dynamical movement primitives: learning attractor models for motor behaviors.
  Neural computation 25~(2), 328--373.

\bibitem[{Islam et~al.(2017)Islam, Henderson, Gomrokchi, and
  Precup}]{islam2017reproducibility}
Islam, R., Henderson, P., Gomrokchi, M., Precup, D., 2017. Reproducibility of
  benchmarked deep reinforcement learning tasks for continuous control. In:
  Proceedings of the ICML 2017 workshop on Reproducibility in Machine Learning
  (RML).

\bibitem[{Jaderberg et~al.(2017)Jaderberg, Dalibard, Osindero, Czarnecki,
  Donahue, Razavi, Vinyals, Green, Dunning, Simonyan,
  et~al.}]{jaderberg2017population}
Jaderberg, M., Dalibard, V., Osindero, S., Czarnecki, W.~M., Donahue, J.,
  Razavi, A., Vinyals, O., Green, T., Dunning, I., Simonyan, K., et~al., 2017.
  Population based training of neural networks. arXiv preprint
  arXiv:1711.09846.

\bibitem[{Jaderberg et~al.(2016)Jaderberg, Mnih, Czarnecki, Schaul, Leibo,
  Silver, and Kavukcuoglu}]{jaderberg2016reinforcement}
Jaderberg, M., Mnih, V., Czarnecki, W.~M., Schaul, T., Leibo, J.~Z., Silver,
  D., Kavukcuoglu, K., 2016. Reinforcement learning with unsupervised auxiliary
  tasks. arXiv preprint arXiv:1611.05397.

\bibitem[{Jonschkowski and Brock(2014)}]{jonschkowski2014state}
Jonschkowski, R., Brock, O., 2014. State representation learning in robotics:
  Using prior knowledge about physical interaction. In: Proceedings of
  Robotics, Science and Systems.

\bibitem[{Kawaguchi(2016)}]{kawaguchi2016deep}
Kawaguchi, K., 2016. Deep learning without poor local minima. In: Advances In
  Neural Information Processing Systems. pp. 586--594.

\bibitem[{Kearns and Singh(2000)}]{kearns2000bias}
Kearns, M.~J., Singh, S.~P., 2000. Bias-variance error bounds for temporal
  difference updates. In: COLT. pp. 142--147.

\bibitem[{Khadka and Tumer(2018)}]{khadka2018evolutionary}
Khadka, S., Tumer, K., 2018. Evolutionary reinforcement learning. arXiv
  preprint arXiv:1805.07917.

\bibitem[{Kober et~al.(2013)Kober, Bagnell, and
  Peters}]{kober2013reinforcement}
Kober, J., Bagnell, J.~A., Peters, J., 2013. Reinforcement learning in
  robotics: A survey. The International Journal of Robotics Research 32~(11),
  1238--1274.

\bibitem[{Kober and Peters(2009)}]{kober2009learning}
Kober, J., Peters, J., 2009. Learning motor primitives for robotics. In: IEEE
  International Conference on Robotics and Automation. IEEE, pp. 2112--2118.

\bibitem[{Koza(1992)}]{koza1992genetic}
Koza, J.~R., 1992. Genetic Programming: On the Programming of Computers by
  Means of Natural Selection. MIT Press, Cambridge, MA.

\bibitem[{Kulkarni et~al.(2016)Kulkarni, Narasimhan, Saeedi, and
  Tenenbaum}]{kulkarni2016hierarchical}
Kulkarni, T.~D., Narasimhan, K.~R., Saeedi, A., Tenenbaum, J.~B., 2016.
  Hierarchical deep reinforcement learning: Integrating temporal abstraction
  and intrinsic motivation. arXiv preprint arXiv:1604.06057.

\bibitem[{Larra{\~n}aga and Lozano(2001)}]{larranaga2001estimation}
Larra{\~n}aga, P., Lozano, J.~A., 2001. Estimation of distribution algorithms:
  A new tool for evolutionary computation. Vol.~2. Springer Science \& Business
  Media.

\bibitem[{Laversanne-Finot et~al.(2018)Laversanne-Finot, P{\'e}r{\'e}, and
  Oudeyer}]{laversanne2018curiosity}
Laversanne-Finot, A., P{\'e}r{\'e}, A., Oudeyer, P.-Y., 2018. Curiosity driven
  exploration of learned disentangled goal spaces. arXiv preprint
  arXiv:1807.01521.

\bibitem[{Lehman et~al.(2017)Lehman, Chen, Clune, and Stanley}]{lehman2017more}
Lehman, J., Chen, J., Clune, J., Stanley, K.~O., 2017. {ES} is more than just a
  traditional finite-difference approximator. arXiv preprint arXiv:1712.06568.

\bibitem[{Lehman and Stanley(2011)}]{lehman2011abandoning}
Lehman, J., Stanley, K.~O., 2011. Abandoning objectives: Evolution through the
  search for novelty alone. Evolutionary computation 19~(2), 189--223.

\bibitem[{Lesort et~al.(2018)Lesort, D{\'\i}az-Rodr{\'\i}guez, Goudou, and
  Filliat}]{lesort2018state}
Lesort, T., D{\'\i}az-Rodr{\'\i}guez, N., Goudou, J.-F., Filliat, D., 2018.
  State representation learning for control: An overview. arXiv preprint
  arXiv:1802.04181.

\bibitem[{Levine and Koltun(2013)}]{levine2013guided}
Levine, S., Koltun, V., 2013. Guided policy search. In: Proceedings of the 30th
  International Conference on Machine Learning. pp. 1--9.

\bibitem[{Levy et~al.(2018)Levy, Platt, and Saenko}]{levy2018hierarchical}
Levy, A., Platt, R., Saenko, K., 2018. Hierarchical reinforcement learning with
  hindsight. arXiv preprint arXiv:1805.08180.

\bibitem[{Lillicrap et~al.(2015)Lillicrap, Hunt, Pritzel, Heess, Erez, Tassa,
  Silver, and Wierstra}]{lillicrap2015continuous}
Lillicrap, T.~P., Hunt, J.~J., Pritzel, A., Heess, N., Erez, T., Tassa, Y.,
  Silver, D., Wierstra, D., 2015. Continuous control with deep reinforcement
  learning. arXiv preprint arXiv:1509.02971.

\bibitem[{Lizotte et~al.(2007)Lizotte, Wang, Bowling, and
  Schuurmans}]{lizotte2007automatic}
Lizotte, D.~J., Wang, T., Bowling, M.~H., Schuurmans, D., 2007. Automatic gait
  optimization with gaussian process regression. In: IJCAI. Vol.~7. pp.
  944--949.

\bibitem[{Mania et~al.(2018)Mania, Guy, and Recht}]{mania2018simple}
Mania, H., Guy, A., Recht, B., 2018. Simple random search provides a
  competitive approach to reinforcement learning. arXiv preprint
  arXiv:1803.07055.

\bibitem[{Martinez-Cantin et~al.(2017)Martinez-Cantin, Tee, and
  McCourt}]{martinez2017policy}
Martinez-Cantin, R., Tee, K., McCourt, M., 2017. Policy search using robust
  {B}ayesian optimization. In: Neural Information Processing Systems (NIPS)
  Workshop on Acting and Interacting in the Real World: Challenges in Robot
  Learning.

\bibitem[{Metzen et~al.(2015)Metzen, Fabisch, and Hansen}]{metzen2015bayesian}
Metzen, J.~H., Fabisch, A., Hansen, J., 2015. Bayesian optimization for
  contextual policy search. In: Proceedings of the Second Machine Learning in
  Planning and Control of Robot Motion Workshop., Hamburg.

\bibitem[{Mnih et~al.(2016)Mnih, Badia, Mirza, Graves, Lillicrap, Harley,
  Silver, and Kavukcuoglu}]{mnih2016asynchronous}
Mnih, V., Badia, A.~P., Mirza, M., Graves, A., Lillicrap, T.~P., Harley, T.,
  Silver, D., Kavukcuoglu, K., 2016. Asynchronous methods for deep
  reinforcement learning. arXiv preprint arXiv:1602.01783.

\bibitem[{Mnih et~al.(2015)Mnih, Kavukcuoglu, Silver, Rusu, Veness, Bellemare,
  Graves, Riedmiller, Fidjeland, Ostrovski, et~al.}]{mnih2015human}
Mnih, V., Kavukcuoglu, K., Silver, D., Rusu, A.~A., Veness, J., Bellemare,
  M.~G., Graves, A., Riedmiller, M., Fidjeland, A.~K., Ostrovski, G., et~al.,
  2015. Human-level control through deep reinforcement learning. Nature
  518~(7540), 529--533.

\bibitem[{Montgomery and Levine(2016)}]{montgomery2016guided}
Montgomery, W.~H., Levine, S., 2016. Guided policy search via approximate
  mirror descent. In: Advances in Neural Information Processing Systems. pp.
  4008--4016.

\bibitem[{Nachum et~al.(2018)Nachum, Gu, Lee, and Levine}]{nachum2018data}
Nachum, O., Gu, S., Lee, H., Levine, S., 2018. Data-efficient hierarchical
  reinforcement learning. arXiv preprint arXiv:1805.08296.

\bibitem[{Najar et~al.(2016)Najar, Sigaud, and Chetouani}]{najar2016training}
Najar, A., Sigaud, O., Chetouani, M., 2016. Training a robot with evaluative
  feedback and unlabeled guidance signals. In: 25th IEEE International
  Symposium on Robot and Human Interactive Communication. IEEE, pp. 261--266.

\bibitem[{Neumann(2011)}]{neumann2011variational}
Neumann, G., 2011. Variational inference for policy search in changing
  situations. In: Proceedings of the 28th international conference on machine
  learning. pp. 817--824.

\bibitem[{O'Donoghue et~al.(2016)O'Donoghue, Munos, Kavukcuoglu, and
  Mnih}]{o2016pgq}
O'Donoghue, B., Munos, R., Kavukcuoglu, K., Mnih, V., 2016. Combining policy
  gradient and q-learning. arXiv preprint arXiv:1611.01626.

\bibitem[{Pelikan et~al.(1999)Pelikan, Goldberg, and
  Cant{\'u}-Paz}]{pelikan1999boa}
Pelikan, M., Goldberg, D.~E., Cant{\'u}-Paz, E., 1999. Boa: The {B}ayesian
  optimization algorithm. In: Proceedings of the 1st Annual Conference on
  Genetic and Evolutionary Computation-Volume 1. Morgan Kaufmann Publishers
  Inc., pp. 525--532.

\bibitem[{Pere et~al.(2018)Pere, Forestier, Sigaud, and
  Oudeyer}]{pere2018unsupervised}
Pere, A., Forestier, S., Sigaud, O., Oudeyer, P.-Y., 2018. Unsupervised
  learning of goal spaces for intrinsically motivated goal exploration. In:
  International Conference on Learning Representations (ICLR). ArXiv preprint
  arXiv:1803.00781.

\bibitem[{Peters et~al.(2010)Peters, M{\"u}lling, and
  Altun}]{peters2010relative}
Peters, J., M{\"u}lling, K., Altun, Y., 2010. Relative entropy policy search.
  In: AAAI. Atlanta, pp. 1607--1612.

\bibitem[{Peters and Schaal(2008{\natexlab{a}})}]{peters08natural}
Peters, J., Schaal, S., 2008{\natexlab{a}}. Natural actor-critic.
  Neurocomputing 71~(7-9), 1180--1190.

\bibitem[{Peters and Schaal(2008{\natexlab{b}})}]{peters2008reinforcement}
Peters, J., Schaal, S., 2008{\natexlab{b}}. Reinforcement learning of motor
  skills with policy gradients. Neural networks 21~(4), 682--697.

\bibitem[{Petroski~Such et~al.(2017)Petroski~Such, Madhavan, Conti, Lehman,
  Stanley, and Clune}]{such2017deep}
Petroski~Such, F., Madhavan, V., Conti, E., Lehman, J., Stanley, K.~O., Clune,
  J., 2017. Deep neuroevolution: Genetic algorithms are a competitive
  alternative for training deep neural networks for reinforcement learning.
  arXiv preprint arXiv:1712.06567.

\bibitem[{Pierrot et~al.(2018)Pierrot, Perrin, and Sigaud}]{pierrot2018first}
Pierrot, T., Perrin, N., Sigaud, O., 2018. First-order and second-order
  variants of the gradient descent: a unified framework. arXiv preprint
  arXiv:1810.08102.

\bibitem[{Plappert et~al.(2017)Plappert, Houthooft, Dhariwal, Sidor, Chen,
  Chen, Asfour, Abbeel, and Andrychowicz}]{plappert2017parameter}
Plappert, M., Houthooft, R., Dhariwal, P., Sidor, S., Chen, R.~Y., Chen, X.,
  Asfour, T., Abbeel, P., Andrychowicz, M., 2017. Parameter space noise for
  exploration. arXiv preprint arXiv:1706.01905.

\bibitem[{Pourchot et~al.(2018)Pourchot, Perrin, and
  Sigaud}]{pourchot2018importance}
Pourchot, A., Perrin, N., Sigaud, O., 2018. Importance mixing: Improving sample
  reuse in evolutionary policy search methods. arXiv preprint arXiv:1808.05832.

\bibitem[{Pourchot and Sigaud(2018)}]{pourchot2018cem}
Pourchot, A., Sigaud, O., 2018. Cem-rl: Combining evolutionary and
  gradient-based methods for policy search. arXiv preprint arXiv:1810.01222.

\bibitem[{Pugh et~al.(2015)Pugh, Soros, Szerlip, and
  Stanley}]{pugh2015confronting}
Pugh, J.~K., Soros, L., Szerlip, P.~A., Stanley, K.~O., 2015. Confronting the
  challenge of quality diversity. In: Proceedings of the 2015 Annual Conference
  on Genetic and Evolutionary Computation. ACM, pp. 967--974.

\bibitem[{Raffin et~al.(2016)Raffin, H{\"o}fer, Jonschkowski, Brock, and
  Stulp}]{raffin2016unsupervised}
Raffin, A., H{\"o}fer, S., Jonschkowski, R., Brock, O., Stulp, F., 2016.
  Unsupervised learning of state representations for multiple tasks. Tech.
  rep., University of Berlin.

\bibitem[{Rastrigin(1963)}]{rastrigin1963convergence}
Rastrigin, L., 1963. The convergence of the random search method in the
  extremal control of a many parameter system. Automation and Remote Control
  24~(10), 1337--1342.

\bibitem[{Riedmiller et~al.(2018)Riedmiller, Hafner, Lampe, Neunert, Degrave,
  Van~de Wiele, Mnih, Heess, and Springenberg}]{riedmiller2018learning}
Riedmiller, M., Hafner, R., Lampe, T., Neunert, M., Degrave, J., Van~de Wiele,
  T., Mnih, V., Heess, N., Springenberg, J.~T., 2018. Learning by
  playing-solving sparse reward tasks from scratch. arXiv preprint
  arXiv:1802.10567.

\bibitem[{Riedmiller et~al.(2008)Riedmiller, Peters, and
  Schaal}]{riedmiller08evaluation}
Riedmiller, M., Peters, J., Schaal, S., 2008. Evaluation of policy gradient
  methods and variants on the cart-pole benchmark. In: {IEEE} International
  Symposium on Approximate Dynamic Programming and Reinforcement Learning
  (ADPRL).

\bibitem[{Rubinstein and Kroese(2004)}]{rubinstein04crossentropy}
Rubinstein, R., Kroese, D., 2004. The Cross-Entropy Method: A Unified Approach
  to Combinatorial Optimization, Monte-Carlo Simulation, and Machine Learning.
  Springer-Verlag.

\bibitem[{Salimans et~al.(2017)Salimans, Ho, Chen, and
  Sutskever}]{salimans2017evolution}
Salimans, T., Ho, J., Chen, X., Sutskever, I., 2017. Evolution strategies as a
  scalable alternative to reinforcement learning. arXiv preprint
  arXiv:1703.03864.

\bibitem[{Schaul et~al.(2015)Schaul, Quan, Antonoglou, and
  Silver}]{schaul2015prioritized}
Schaul, T., Quan, J., Antonoglou, I., Silver, D., 2015. Prioritized experience
  replay. arXiv preprint arXiv:1511.05952.

\bibitem[{Schulman et~al.(2015)Schulman, Levine, Moritz, Jordan, and
  Abbeel}]{schulman2015trust}
Schulman, J., Levine, S., Moritz, P., Jordan, M.~I., Abbeel, P., 2015. Trust
  region policy optimization. CoRR, abs/1502.05477.

\bibitem[{Schulman et~al.(2017)Schulman, Wolski, Dhariwal, Radford, and
  Klimov}]{schulman2017proximal}
Schulman, J., Wolski, F., Dhariwal, P., Radford, A., Klimov, O., 2017. Proximal
  policy optimization algorithms. arXiv preprint arXiv:1707.06347.

\bibitem[{Sehnke et~al.(2010)Sehnke, Osendorfer, R{\"u}ckstie{\ss}, Graves,
  Peters, and Schmidhuber}]{sehnke2010parameter}
Sehnke, F., Osendorfer, C., R{\"u}ckstie{\ss}, T., Graves, A., Peters, J.,
  Schmidhuber, J., 2010. Parameter-exploring policy gradients. Neural Networks
  23~(4), 551--559.

\bibitem[{Shelhamer et~al.(2016)Shelhamer, Mahmoudieh, Argus, and
  Darrell}]{shelhamer2016loss}
Shelhamer, E., Mahmoudieh, P., Argus, M., Darrell, T., 2016. Loss is its own
  reward: Self-supervision for reinforcement learning. arXiv preprint
  arXiv:1612.07307.

\bibitem[{Sigaud and Buffet(2010)}]{sigaud2010markov}
Sigaud, O., Buffet, O., 2010. Markov Decision Processes in Artificial
  Intelligence. iSTE - Wiley.

\bibitem[{Silver et~al.(2014)Silver, Lever, Heess, Degris, Wierstra, and
  Riedmiller}]{silver2014deterministic}
Silver, D., Lever, G., Heess, N., Degris, T., Wierstra, D., Riedmiller, M.,
  2014. Deterministic policy gradient algorithms. In: Proceedings of the 30th
  International Conference in Machine Learning.

\bibitem[{Stanley and Miikkulainen(2002)}]{stanley2002efficient}
Stanley, K.~O., Miikkulainen, R., 2002. Efficient evolution of neural network
  topologies. In: Evolutionary Computation, 2002. CEC'02. Proceedings of the
  2002 Congress on. Vol.~2. IEEE, pp. 1757--1762.

\bibitem[{Stulp and Sigaud(2012{\natexlab{a}})}]{stulp12icml}
Stulp, F., Sigaud, O., 2012{\natexlab{a}}. Path integral policy improvement
  with covariance matrix adaptation. In: Proceedings of the 29th
  {I}nternational {C}onference on {M}achine {L}earning. Edinburgh, Scotland,
  pp. 1--8.

\bibitem[{Stulp and Sigaud(2012{\natexlab{b}})}]{stulp2012policy}
Stulp, F., Sigaud, O., 2012{\natexlab{b}}. Policy improvement methods: Between
  black-box optimization and episodic reinforcement learning. Tech. rep.,
  hal-00738463.

\bibitem[{Stulp and Sigaud(2013)}]{stulp13paladyn}
Stulp, F., Sigaud, O., august 2013. Robot skill learning: From reinforcement
  learning to evolution strategies. Paladyn Journal of Behavioral Robotics
  4~(1), 49--61.

\bibitem[{Stulp and Sigaud(2015)}]{stulp15NN}
Stulp, F., Sigaud, O., 2015. Many regression algorithms, one unified model: A
  review. Neural Networks 69, 60--79.

\bibitem[{Sun et~al.(2009)Sun, Wierstra, Schaul, and
  Schmidhuber}]{sun2009efficient}
Sun, Y., Wierstra, D., Schaul, T., Schmidhuber, J., 2009. Efficient natural
  evolution strategies. In: Proceedings of the 11th Annual conference on
  Genetic and evolutionary computation. ACM, pp. 539--546.

\bibitem[{Sutton(1988)}]{sutton88}
Sutton, R.~S., 1988. {L}earning to {P}redict by the {M}ethod of {T}emporal
  {D}ifferences. Machine Learning 3, 9--44.

\bibitem[{Sutton and Barto(1998)}]{sutton98}
Sutton, R.~S., Barto, A.~G., 1998. Reinforcement {L}earning: An {I}ntroduction.
  MIT Press.

\bibitem[{Tang and Kucukelbir(2017)}]{tang2017variational}
Tang, Y., Kucukelbir, A., 2017. Variational deep q network. arXiv preprint
  arXiv:1711.11225.

\bibitem[{Theodorou et~al.(2010)Theodorou, Buchli, and
  Schaal}]{theodorou10generalized}
Theodorou, E., Buchli, J., Schaal, S., 2010. A generalized path integral
  control approach to reinforcement learning. Journal of Machine Learning
  Research 11, 3137--3181.

\bibitem[{Thrun and Mitchell(1995)}]{thrun1995lifelong}
Thrun, S., Mitchell, T.~M., 1995. Lifelong robot learning. Robotics and
  autonomous systems 15~(1-2), 25--46.

\bibitem[{Togelius et~al.(2009)Togelius, Schaul, Wierstra, Igel, Gomez, and
  Schmidhuber}]{togelius2009ontogenetic}
Togelius, J., Schaul, T., Wierstra, D., Igel, C., Gomez, F., Schmidhuber, J.,
  2009. Ontogenetic and phylogenetic reinforcement learning. K{\"u}nstliche
  Intelligenz 23~(3), 30--33.

\bibitem[{Veeriah et~al.(2018)Veeriah, Oh, and Singh}]{veeriah2018many}
Veeriah, V., Oh, J., Singh, S., 2018. Many-goals reinforcement learning. arXiv
  preprint arXiv:1806.09605.

\bibitem[{Vezhnevets et~al.(2017)Vezhnevets, Osindero, Schaul, Heess,
  Jaderberg, Silver, and Kavukcuoglu}]{vezhnevets2017feudal}
Vezhnevets, A.~S., Osindero, S., Schaul, T., Heess, N., Jaderberg, M., Silver,
  D., Kavukcuoglu, K., 2017. Feudal networks for hierarchical reinforcement
  learning. arXiv preprint arXiv:1703.01161.

\bibitem[{Wang et~al.(2016{\natexlab{a}})Wang, Kurth-Nelson, Tirumala, Soyer,
  Leibo, Munos, Blundell, Kumaran, and Botvinick}]{wang2016learning}
Wang, J.~X., Kurth-Nelson, Z., Tirumala, D., Soyer, H., Leibo, J.~Z., Munos,
  R., Blundell, C., Kumaran, D., Botvinick, M., 2016{\natexlab{a}}. Learning to
  reinforcement learn. arXiv preprint arXiv:1611.05763.

\bibitem[{Wang et~al.(2016{\natexlab{b}})Wang, Bapst, Heess, Mnih, Munos,
  Kavukcuoglu, and de~Freitas}]{wang2016sample}
Wang, Z., Bapst, V., Heess, N., Mnih, V., Munos, R., Kavukcuoglu, K.,
  de~Freitas, N., 2016{\natexlab{b}}. Sample efficient actor-critic with
  experience replay. arXiv preprint arXiv:1611.01224.

\bibitem[{Wierstra et~al.(2008)Wierstra, Schaul, Peters, and
  Schmidhuber}]{wierstra2008natural}
Wierstra, D., Schaul, T., Peters, J., Schmidhuber, J., 2008. Natural evolution
  strategies. In: IEEE Congress on Evolutionary Computation. IEEE, pp.
  3381--3387.

\bibitem[{Williams and Singh(1998)}]{williams1998experimental}
Williams, J.~K., Singh, S.~P., 1998. Experimental results on learning
  stochastic memoryless policies for partially observable markov decision
  processes. In: NIPS. pp. 1073--1080.

\bibitem[{Williams(1992)}]{williams1992}
Williams, R.~J., May 1992. {Simple statistical gradient-following algorithms
  for connectionist reinforcement learning}. Machine Learning 8~(3-4),
  229--256.

\bibitem[{Wilson et~al.(2014)Wilson, Fern, and Tadepalli}]{wilson2014using}
Wilson, A., Fern, A., Tadepalli, P., 2014. Using trajectory data to improve
  bayesian optimization for reinforcement learning. The Journal of Machine
  Learning Research 15~(1), 253--282.

\bibitem[{Wu et~al.(2017)Wu, Mansimov, Liao, Grosse, and Ba}]{wu2017scalable}
Wu, Y., Mansimov, E., Liao, S., Grosse, R., Ba, J., 2017. Scalable trust-region
  method for deep reinforcement learning using {K}ronecker-factored
  approximation. arXiv preprint arXiv:1708.05144.

\bibitem[{Yang and Hospedales(2014)}]{yang2014unified}
Yang, Y., Hospedales, T.~M., 2014. A unified perspective on multi-domain and
  multi-task learning. arXiv preprint arXiv:1412.7489.

\bibitem[{Zhang et~al.(2017)Zhang, Clune, and Stanley}]{zhang2017relationship}
Zhang, X., Clune, J., Stanley, K.~O., 2017. On the relationship between the
  openai evolution strategy and stochastic gradient descent. arXiv preprint
  arXiv:1712.06564.

\bibitem[{Zimmer and Doncieux(2017)}]{zimmer2017bootstrapping}
Zimmer, M., Doncieux, S., 2017. Bootstrapping {Q}-learning for robotics from
  neuro-evolution results. IEEE Transactions on Cognitive and Developmental
  Systems.

\end{thebibliography}

\end{document}